\documentclass[letterpaper, 10 pt, conference]{ieeeconf}  

\IEEEoverridecommandlockouts                              

\usepackage{graphicx} 
\usepackage[caption=false,font=footnotesize]{subfig}
\usepackage{amsmath} 
\usepackage{amssymb}  
\usepackage{booktabs}
\usepackage{makecell}
\usepackage{comment}
\usepackage{multirow}
\usepackage{pifont}

\usepackage{flushend}
\let\labelindent\relax
\usepackage{enumitem}
\usepackage{hyperref}

\title{\LARGE \bf
SURE: \underline{S}afe \underline{U}ncertainty-Aware \underline{R}obot-\underline{E}nvironment Interaction using Trajectory Optimization
}

\author{\authorblockN{Zhuocheng Zhang\authorrefmark{1}, Haizhou Zhao\authorrefmark{1}\authorrefmark{2}, Xudong Sun\authorrefmark{1}, Aaron M. Johnson\authorrefmark{1}\authorrefmark{3}, Majid Khadiv\authorrefmark{1}}
\authorblockA{\authorrefmark{1}Technical University of Munich (TUM), Germany. \quad \authorrefmark{2}New York University (NYU), USA.}
\authorblockA{\authorrefmark{3}Carnegie Mellon University (CMU), USA.}}

\begin{document}

\maketitle
\thispagestyle{empty}
\pagestyle{empty}

\begin{abstract}
Robotic tasks involving contact interactions pose significant challenges for Trajectory Optimization (TO) due to discontinuous dynamics. Conventional formulations typically assume deterministic contact events, limiting robustness and adaptability in real-world settings. In this work, we propose SURE, a robust TO framework that explicitly accounts for uncertainty in contact timing.
By allowing multiple trajectories to branch from possible pre-impact states and subsequently merge into a shared trajectory, SURE achieves both robustness and computational efficiency within a unified optimization framework.
We evaluate SURE on two representative tasks with unknown impact times. In a cart-pole impact task with uncertain wall location, SURE achieves an average improvement of 21.6\% in success rate when branch switching is enabled during control. In an egg-catching experiment using a robotic manipulator, SURE improves the success rate by 40\%.
These results demonstrate that SURE substantially enhances robustness compared to conventional nominal formulations. A supplementary video demonstrating the results is available \href{https://www.youtube.com/watch?v=EhXzrgQxk3g}{here}.
\end{abstract}

\section{INTRODUCTION}
Locomotion and manipulation tasks inherently involve intermittent contact with the environment, leading to discrete switches in system dynamics and resulting in hybrid dynamical systems~\cite{johnson2016hybrid}. Trajectory Optimization (TO) for such systems is typically formulated as either a mathematical program with complementarity constraints (MPCC)~\cite{scheel2000mathematical, patel2019contact, kim2025contact, aydinoglu2024consensus} or a mixed-integer optimization problem~\cite{deits2014footstep, aceituno2017simultaneous, ponton2021efficient}.
Most recent TO approaches for contact-rich problems adopt hierarchical formulations~\cite{wensing2023optimization}, separating contact planning from whole-body motion generation. These methods typically assume deterministic contact timings~\cite{grandia2023perceptive, dhedin2025simultaneous}. In practice, however, this assumption is restrictive: environmental perception is inherently uncertain, and modeling errors or end-effector tracking inaccuracies during contact establishment can lead to discrepancies between planned and actual contact times. 

The main contribution of this work is the development of a robust TO framework, termed SURE, with uncertain contact timing. The framework allows trajectories to split from possible pre-impact states and later rejoin a shared trajectory, thereby achieving both robustness and computational efficiency.
Through two case studies, we demonstrate improved robustness over nominal TO.

The remainder of this paper is organized as follows.
Section~\ref{sec:related_work} reviews related work, Section~\ref{sec:method} introduces the nominal TO formulation, and presents the SURE formulation and its applications. Section~\ref{sec:results} reports the experimental results, and Section~\ref{sec:conclusion} concludes the paper and outlines future directions.

\begin{figure}[t!]
	\centering
	\includegraphics[width=.47\textwidth]{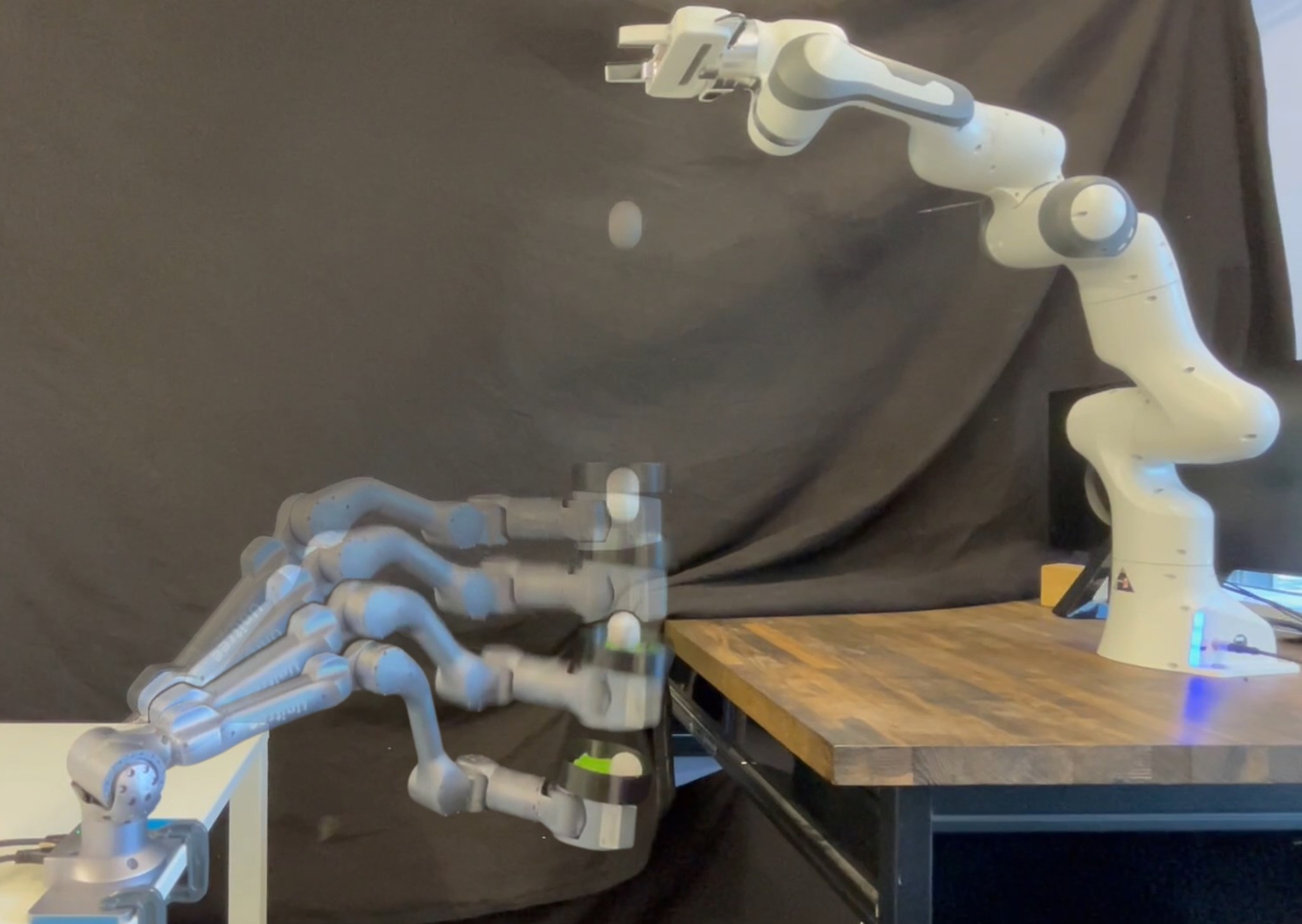}
	\caption{Illustration of the egg-catching task. The Unitree Z1 robot arm attempts to catch a falling egg while minimizing impact. In this task, millisecond-level timing deviations can significantly affect performance, highlighting the importance of robustness to contact-timing uncertainty.}
	\label{fig:egg_catching_demo}
\end{figure}

\section{RELATED WORK}
\label{sec:related_work}
Robust and stochastic optimal control methods have been widely studied for contact-rich robotic systems~\cite{hammoud2021impedance, gazar2023multi, shirai2024chance, drnach2021robust}. However, they typically address contact timing uncertainty only at the feedback level, without adapting the nominal trajectory, which limits performance in scenarios such as delayed contact events.
Some approaches constrain predicted post-impact states~\cite{wang2023impact} or introduce reference spreading~\cite{van2024quadratic}, but their lack of a prediction horizon can lead to myopic behavior. Others leverage the saltation matrix to model hybrid transitions~\cite{zhu2022hybrid, zhu2024convergent}, yet rely on local linearization and do not account for control adaptation under timing mismatch. 
Some methods partially address these issues:~\cite{green2020planning} optimizes trajectories across disturbances with shared controls, while~\cite{zhao2024trajectory} proposes a robust formulation for timing uncertainty. However, these approaches do not capture long-term optimality or dependencies on post-impact events.

The closest work to ours is \cite{zhao2024trajectory}, where the authors proposed a trajectory optimization formulation that is robust to contact timing uncertainties. However, unlike the nominal formulation (Fig.~\ref{fig:nominaltrajopt_sketch}), this approach cannot explicitly generate a trajectory that reaches the terminal state (Fig.~\ref{fig:robusttrajopt_sketch}). Instead, the trajectory terminates at the latest possible contact time, while the subsequent motion is handled through instantaneous replanning. Intuitively, this formulation aims to keep the post-contact states favorable, but it does not provide an explicit plan for reaching the terminal state. In addition to this theoretical improvement, we present the first experimental validation of such a robust formulation.

Another possible approach to handling contact timing uncertainty is to adopt the tree-structured topology underlying the Tree Optimal Control Problem (OCP) formulation proposed in \cite{frison2020hpipm}. Although this framework was originally developed for solving Quadratic Programs, we use only its topology as a structural reference. With suitable pruning, it can be adapted to a more general nonlinear trajectory optimization problem considered in this work, where contact-timing uncertainty is handled by enumerating trajectories that branch from each possible contact node and propagate to the terminal state, as illustrated in Fig.~\ref{fig:tree_ocp_qp}. However, this brute-force strategy increases the computational complexity of the trajectory optimization problem, making it impractical for high-dimensional systems.
Motivated by these limitations, our goal in this work is to develop a robust trajectory optimization framework that accounts for uncertainty while avoiding excessive computational complexity.

In reinforcement learning (RL), several works address uncertainty-aware control. \cite{bogdanovic2020learning,khader2020stability} introduce variable-impedance controllers for robustness to contact variations, while \cite{zhi2025learning,xu2025facet} learn unified position–force policies to handle unexpected external forces. \cite{kuo2021uncertainty} incorporates Gaussian Process dynamics within model-based RL, and recent approaches rely on domain randomization to improve robustness in contact-rich manipulation \cite{barreiros2025learning}. However, RL methods typically lack hard constraints on internal variables such as impact forces, raising safety concerns.
Despite this, RL insights can inform more robust optimization-based control frameworks.

\section{METHOD}\label{sec:method}
\subsection{Nominal Trajectory Optimization}
\label{sec:nominal}
\begin{figure}[t!]
  \centering
  \subfloat[Nominal Approach]{\includegraphics[width=0.48\linewidth]{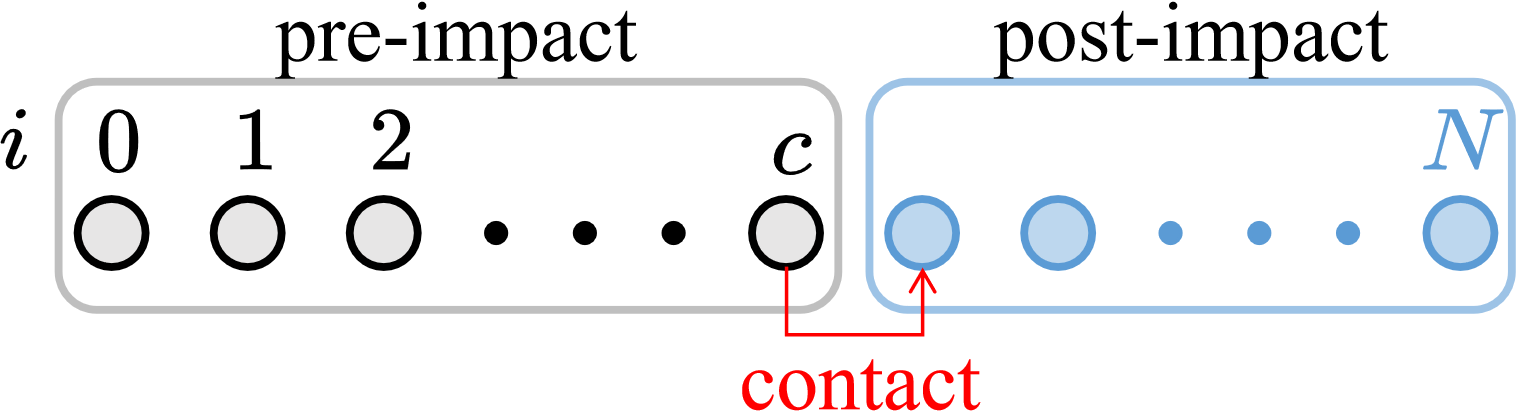}\label{fig:nominaltrajopt_sketch}}
  \hfill
  \subfloat[Approach in \cite{zhao2024trajectory}]{\includegraphics[width=0.48\linewidth]{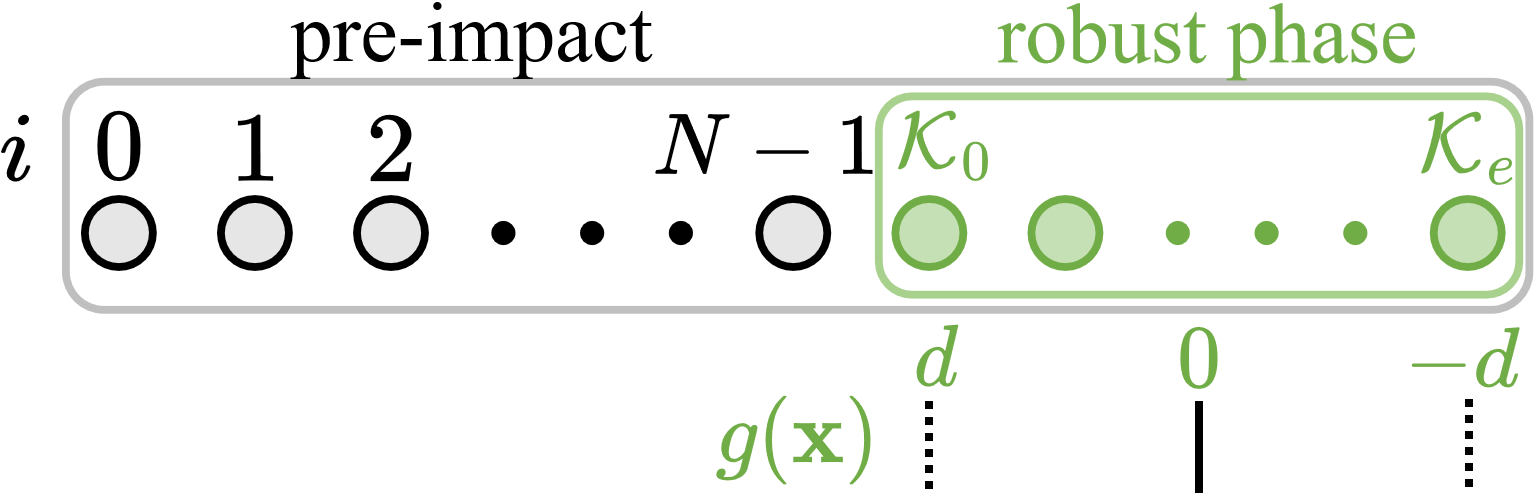}\label{fig:robusttrajopt_sketch}}
  \hfill
  \subfloat[Tree OCP\cite{frison2020hpipm} \& its pruning]{\includegraphics[width=0.48\linewidth]{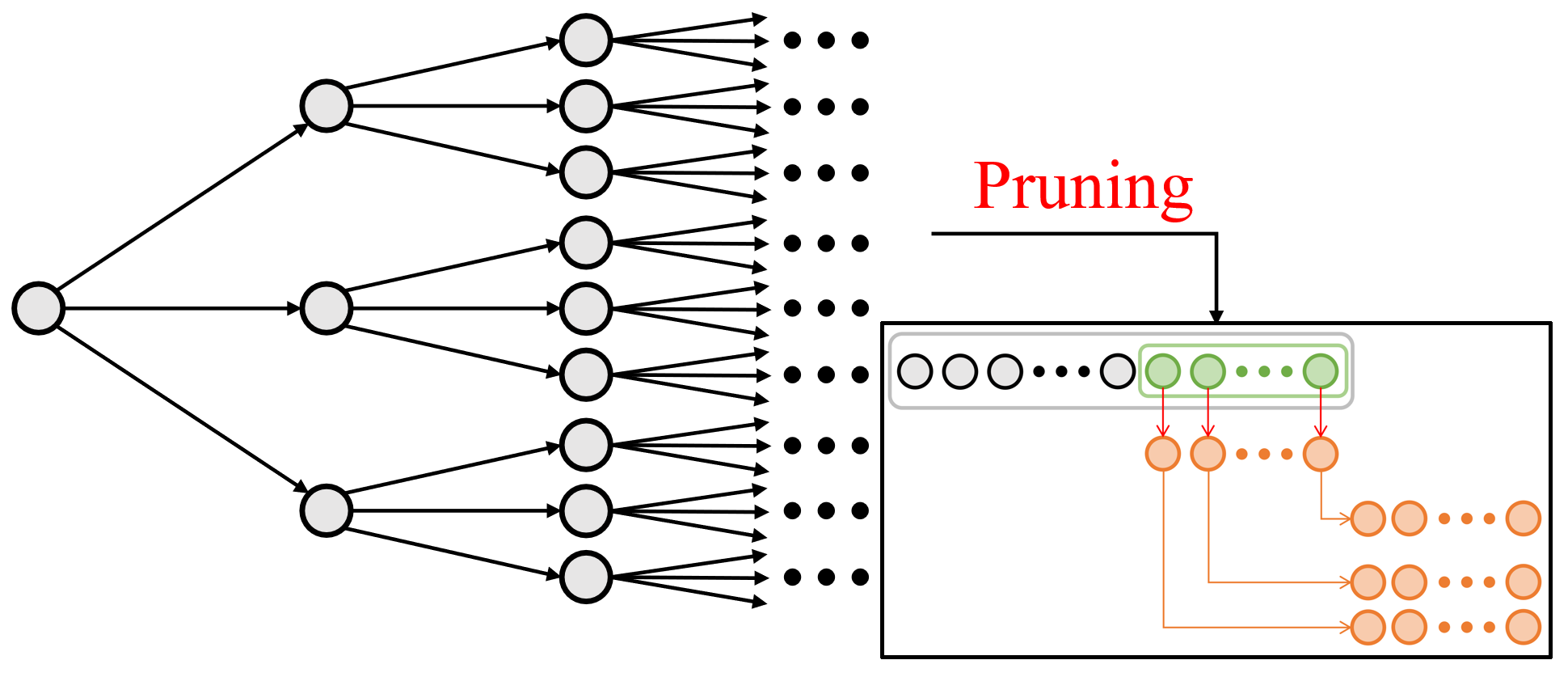}\label{fig:tree_ocp_qp}}
  \hfill
  \subfloat[SURE (ours)]{\includegraphics[width=0.48\linewidth]{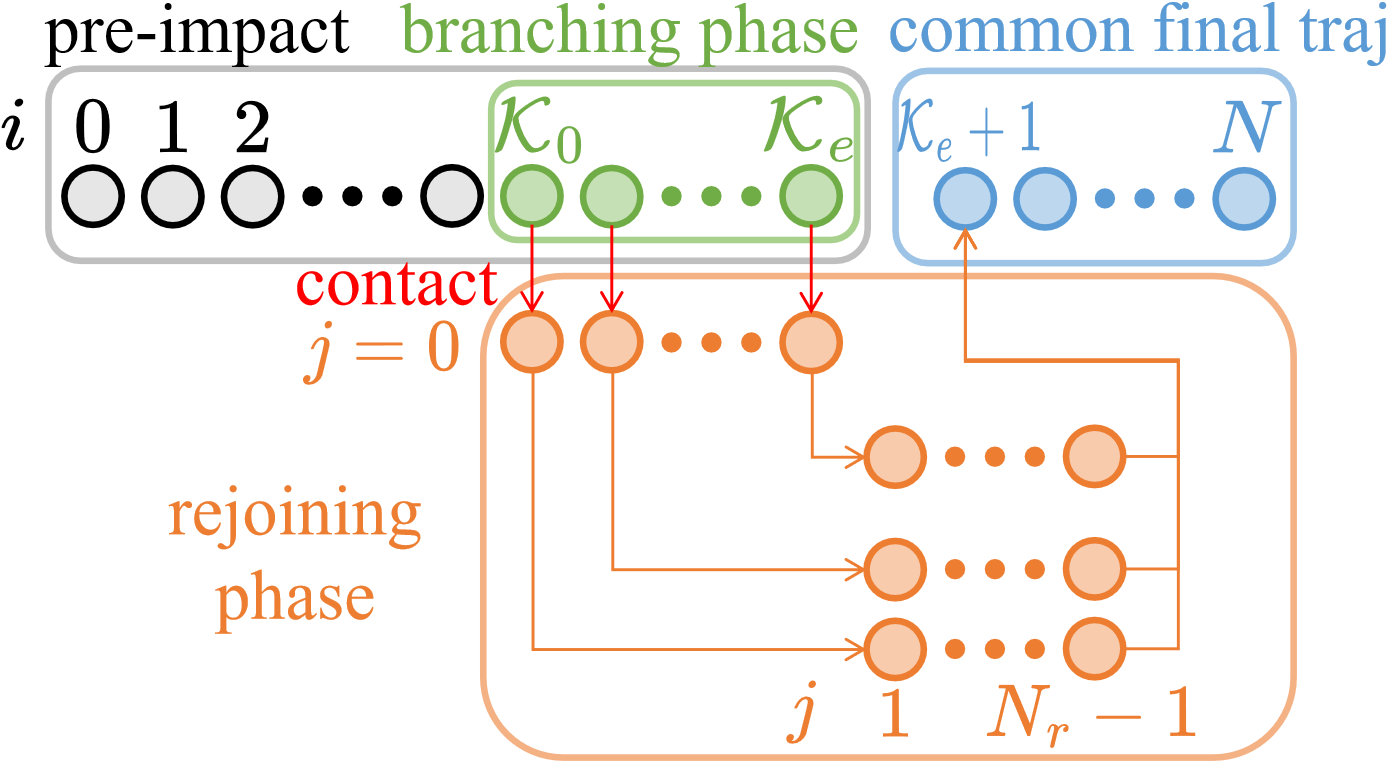}\label{fig:rejoiningtrajopt_sketch}}
  \caption{(a) Nominal trajectory optimization, where contact occurs at node $c$, separating pre- and post-impact phases. (b) Method in~\cite{zhao2024trajectory}, allowing contact at any node within a robust phase. (c) Tree OCP~\cite{frison2020hpipm} formulation topology and our pruned adaptation for solving the contact-timing uncertainty problem, which requires many nodes and incurs high computational cost due to brute-force enumeration. (d) SURE: gray nodes denote the pre-contact trajectory, green nodes the branching phase (as in (b)), and blue nodes the common final trajectory. Together they form the shared trajectory. Orange nodes represent branch trajectories that later rejoin the common trajectory.}\label{fig:comp_phase_diagram_nominal_robust}
\end{figure}

In this section, we present the formulation of the nominal TO using multiple shooting, which assumes that contact occurs at a deterministic time.
The trajectory starts from the initial state $\mathbf{x}_0$, under control $\mathbf{u}$, and is discretized into $N$ shooting nodes, Fig.~\ref{fig:nominaltrajopt_sketch}. We assume that the contact occurs at the node indexed by $c$. The TO problem can then be formulated as follows:
\begin{subequations}\label{eq:nominaltrajopt}
\begin{align}
\min_{\mathbf{x},\mathbf{u},\Delta t}\quad&\sum_{i=0}^{N-1}L_i(\mathbf{x}_i, \mathbf{u}_i, \Delta t_{i})+L_{N}(\mathbf{x}_N)\\
\text{s.t.}\quad&\forall i\in[0, N]: \hspace{0.4cm}\mathbf{w}_i(\mathbf{x}_i, \mathbf{u}_i)=0,\label{eq:nominaltrajopt_runningeqcons}\\
    &\hspace{2.3cm}\mathbf{h}_i(\mathbf{x}_i, \mathbf{u}_i)\leq 0,\label{eq:nominaltrajopt_runningueqcons}\\
&\forall i \notin \{c, N\}:  \hspace{0.3cm}\mathbf{f}_i(\mathbf{x}_i, \mathbf{u}_i, \mathbf{x}_{i+1}, \Delta t_i)=0,\label{eq:nominaltrajopt_dynamics}\\
    &\hspace{2.3cm} g(\mathbf{x}_i)>0,\label{eq:nominaltrajopt_guardueq}\\
&i=c: \hspace{1.3cm}
\mathbf{x}_{i+1} = R(\mathbf{x}_{i}),\label{eq:nominaltrajopt_resetmap}\\
& \hspace{2.3cm} g(\mathbf{x}_i) = 0\label{eq:nominaltrajopt_guardeq},\\
&\Delta t\in[\Delta t_{\min},\Delta t_{\max}].\label{eq:nominaltrajopt_timebound}
\end{align}
\end{subequations}
Here, $\Delta t_{i}$ denotes the time interval between nodes, varying within the bounds specified in ~\eqref{eq:nominaltrajopt_timebound}; $L_N$ is the terminal cost; $L_i$ is the running cost; $\mathbf{w}$ and $\mathbf{h}$ represent the state-input equality and inequality constraints, respectively; $\mathbf{f}$ specifies the discrete dynamics; $g$ is the guard function, whose value determines the triggering condition of the contact event (contact occurs at $g=0$, while $g>0$ holds at all other times); and $R$ is the impact or reset function.

\subsection{SURE Formulation}\label{sec:sure}
Similar to the robust phase in~\cite{zhao2024trajectory}, we construct a branching phase $\mathcal{K}$, which includes all the possible pre-impact nodes. To address the drawback of~\cite{zhao2024trajectory}, we force these branches to rejoin a common final trajectory. From each node in the branching phase, we generate a rejoining phase trajectory, each of which ends at the start of this common final trajectory, as shown in Fig.~\ref{fig:rejoiningtrajopt_sketch}. For trajectories with multiple contact transitions, this process is repeated for each.

In this way, uncertainty is explicitly accounted for, while avoiding the need to plan separate trajectories from each post-impact state to the terminal state. Instead, all branches rejoin a common state, from which they evolve jointly toward the terminal state. This strategy trades a degree of optimality for substantially improved computational efficiency, as it greatly reduces the number of decision variables.

The SURE formulation under uncertain contact timing is formulated as follows:
\begin{subequations}\label{eq:rejoiningtrajopt}
\allowdisplaybreaks
\begin{align}
\min_{\mathbf{x}, \mathbf{u},\Delta t,d}&\sum_{i=0}^{N-1}L_i(\mathbf{x}_i, \mathbf{u}_i, \Delta t_{i}) + \sum_{i\in\mathcal{K}}\sum_{j=0}^{N_r}L_{i,j}(\mathbf{x}_{i,j}, \mathbf{u}_{i,j}, \Delta t_{i,j})\nonumber
\\&\qquad+L_{N}(\mathbf{x}_N)\label{eq:rejoiningtrajopt_cost}\\
\text{s.t.}\quad&\forall i\in[0,N]:\hspace{0.2cm}
    \mathbf{w}_i(\mathbf{x}_i, \mathbf{u}_i)=0,\label{eq:rejoiningtrajopt_commoneqcons}\\
    &\hspace{2.1cm}\mathbf{h}_i(\mathbf{x}_i, \mathbf{u}_i)\leq 0,\label{eq:rejoiningtrajopt_commonueqcons}\\
&\forall i \notin \{\mathcal{K}_e,N\}:
    \mathbf{f}_i(\mathbf{x}_i,\mathbf{u}_i,\mathbf{x}_{i+1},\Delta t_i)=0,\\
&\forall i\in\mathcal{K}:\hspace{0.8cm}
    \mathbf{x}_{i,0}=R(\mathbf{x}_i),\label{eq:rejoiningtrajopt_robustnodesreset}\\
    &\hspace{2.1cm}\mathbf{x}_{i,N_r}=\mathbf{x}_{\mathcal{K}_{e}+1},\label{eq:rejoiningtrajopt_branchterminalcons}\\
    &\qquad\forall j\in[0,N_r-1]:\nonumber\\
        &\hspace{2.1cm}\mathbf{f}_{i,j}(\mathbf{x}_{i,j},\mathbf{u}_{i,j},\mathbf{x}_{i,j+1},\Delta t_{i,j})=0,\label{eq:rejoiningtrajopt_branchdynamics}\\
        &\hspace{2.1cm}\mathbf{w}_{i,j}(\mathbf{x}_{i,j}, \mathbf{u}_{i,j})=0,\label{eq:rejoiningtrajopt_brancheqcons}\\
        &\hspace{2.1cm}\mathbf{h}_{i,j}(\mathbf{x}_{i,j}, \mathbf{u}_{i,j})\leq 0,\label{eq:rejoiningtrajopt_branchueqcons}\\
&i=\mathcal{K}_0:\hspace{0.8cm} g(\mathbf{x}_i)=d,\label{eq:rejoiningtrajopt_firstguard}\\
&i=\mathcal{K}_{e}:\hspace{0.8cm} g(\mathbf{x}_i)=-d,\label{eq:rejoiningtrajopt_lastguard}\\
&\forall i<\mathcal{K}_0:\hspace{0.6cm}
     g(\mathbf{x}_i)>d,\label{eq:rejoiningtrajopt_beyondguardcondition}\\
&d\in[d_{\min},d_{\max}],\label{eq:rejoiningtrajopt_uncertaintybound}\\
&\Delta t\in[\Delta t_{\min},\Delta t_{\max}].
\end{align}
\end{subequations}

All nodes can be classified into two categories: (1) nodes on the common trajectory (with a single subscript index, e.g., $\mathbf{x}_i$) and (2) nodes in the rejoining phase (with two subscript indices, e.g., $\mathbf{x}_{i,j}$). 

On the common trajectory, $\mathcal{K}_0$ and $\mathcal{K}_{e}$ denote the indices of the first and last nodes in the branching phase, respectively. The variable $d$ denotes half the width of the uncertainty range. At $i = \mathcal{K}_0$, the earliest possible contact, we require $g(\mathbf{x}_i) = d$, \eqref{eq:rejoiningtrajopt_firstguard}, while at $i = \mathcal{K}_{e}$, the latest possible contact, we require $g(\mathbf{x}_i) = -d$, \eqref{eq:rejoiningtrajopt_lastguard}. Each node in the branching phase represents a pre-contact state, and we apply a state transition to compute the corresponding post-impact state which is then the initial state on the corresponding branch,\eqref{eq:rejoiningtrajopt_robustnodesreset}. 

In the rejoining phase, each branch contains $N_r + 1$ nodes, indexed by $j = 0, 1, \ldots, N_r$. Node $\mathbf{x}_{i,j}$ denotes the $j$-th node on the branch originating from node $i \in \mathcal{K}$. The contact-related cost and constraints are incorporated into $L_{i,j}$ as well as~\eqref{eq:rejoiningtrajopt_brancheqcons} and~\eqref{eq:rejoiningtrajopt_branchueqcons}. At $j = N_r$, the branch node rejoins the common final trajectory at the node with index $i=\mathcal{K}_{e}+1$, \eqref{eq:rejoiningtrajopt_branchterminalcons}. 
In practice, the half-width of the uncertainty range, $d$, can be set as a fixed but tunable parameter, or $d$ can also be a decision variable and incorporated into the cost function. In that case, the solver aims to identify the largest possible uncertainty region over which the system remains robust.

Another parameter that influences robustness is the number of branches, $|\mathcal{K}|$. On the one hand, for the sake of computational efficiency, it is desirable to keep the number of decision variables as small as possible. On the other hand, increasing the number of branches generally improves the robustness of the resulting trajectory. Therefore, a proper trade-off must be found. 

\subsection{Application of Solution Trajectories for Control}
\label{sec:sure_application}
Solving~\eqref{eq:rejoiningtrajopt} yields a family of trajectories for different contact timings. During control, these can serve as references. Depending on the availability of contact sensing, two approaches can be used to exploit it.
\begin{figure}[!t]
  \vspace*{5pt}
  \centering
  \subfloat[Trajectory scheduling]{\includegraphics[width=0.45\linewidth]{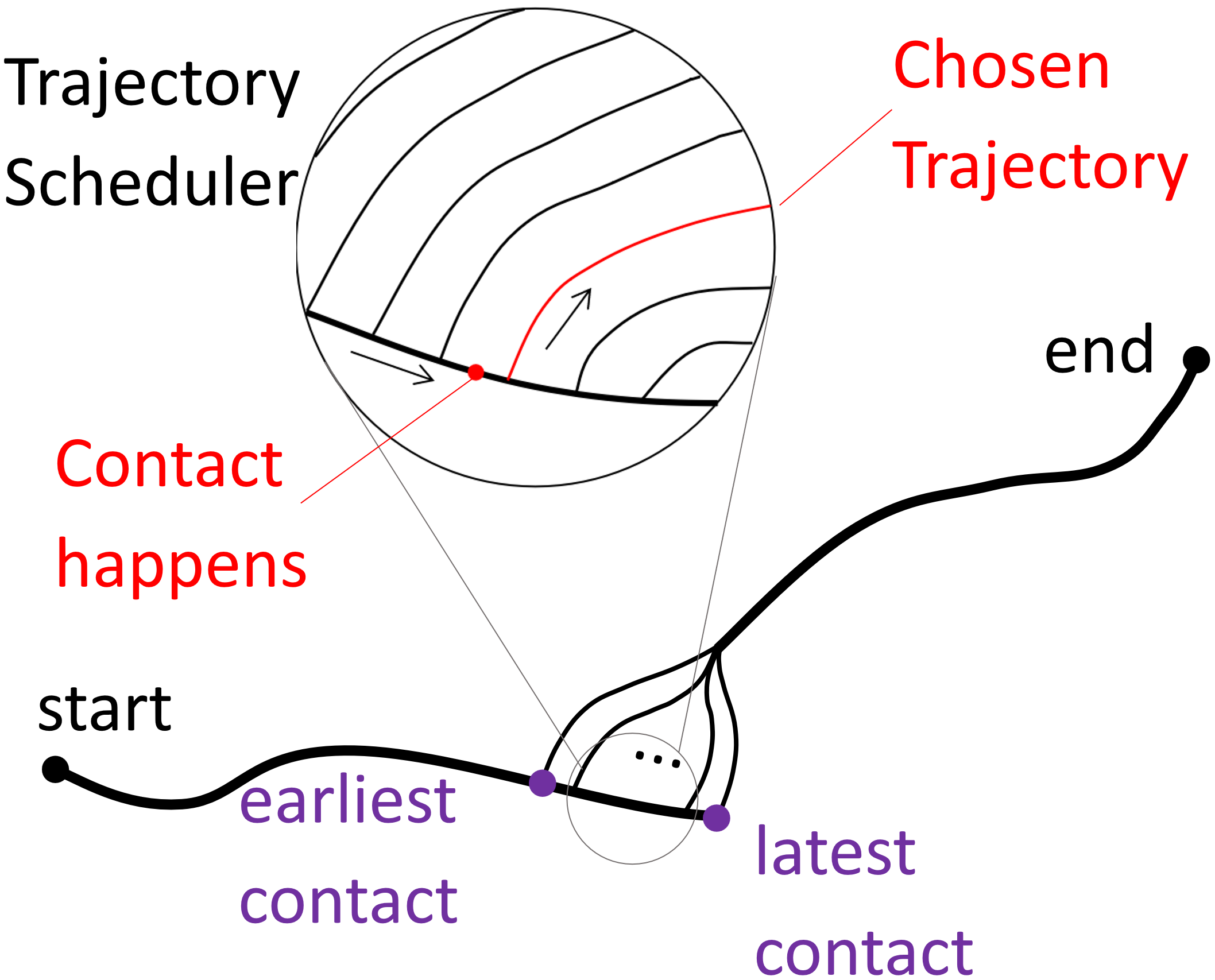}\label{fig:trajectory_scheduling}}
  \subfloat[Robust nominal trajectory]{\includegraphics[width=0.45\linewidth]{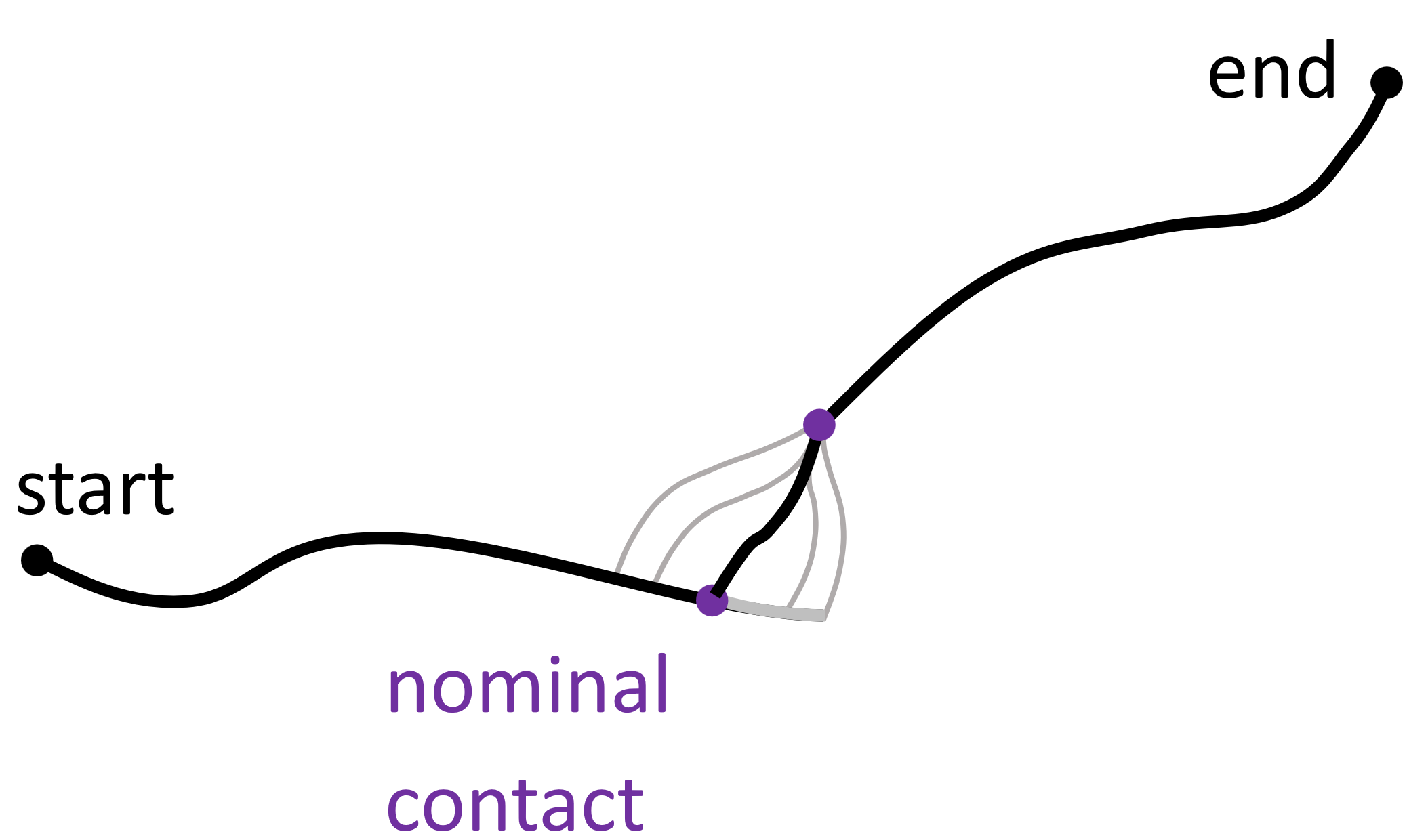}\label{fig:robust_nominal_traj}}
  \caption{Illustrations of (a) trajectory scheduling and (b) the robust nominal trajectory. Trajectory scheduling uses all optimized trajectories from the robust formulation, while the robust nominal trajectory corresponds to the middle branch among them.}\label{fig:illustration_application_rejoining_trajs}
\end{figure}

\subsubsection{Trajectory Scheduling}\label{sec:trajectory_scheduling}
When contact sensing is available, the reference can be switched based on the detected contact time, referred to as \emph{trajectory scheduling} (Fig.~\ref{fig:trajectory_scheduling}). The system initially follows the common pre-impact trajectory. During the branching phase, it continues along the common trajectory if no contact occurs. Upon contact, the scheduler switches to the nearest subsequent trajectory, guiding the system to the terminal state.

\subsubsection{Robust Nominal Trajectory}\label{sec:rejoining_nominal_trajectory}
If contact sensors are not available, multiple trajectories remain beneficial. By introducing additional decision variables and constraints, the pre-contact trajectory can be reshaped to reduce worst-case impact within the uncertainty region. As a result, selecting a branch is expected to improve worst-case performance compared to the nominal trajectory, although performance in specific cases may degrade.
A natural choice is the middle branch, as it lies approximately equidistant from the extremes of the uncertainty set and is therefore most representative. This trajectory branches at index $i=\lceil (\mathcal{K}_0+\mathcal{K}_{e})/2 \rceil$ and is referred to as the \emph{robust nominal trajectory} (Fig.~\ref{fig:robust_nominal_traj}).

\section{Case Studies}\label{sec:results}
In this section, we present two case studies to evaluate trajectories from SURE and compare them with nominal formulations. All problems are implemented in CasADi~\cite{andersson2019casadi} using the Opti stack and solved as nonlinear programs with IPOPT~\cite{wachter2006implementation}.
\subsection{Case Study I: Cart-Pole System with Wall}
\label{sec:casestudy1}
\begin{figure}[t!]
	\centering
	\includegraphics[width=.3\textwidth]{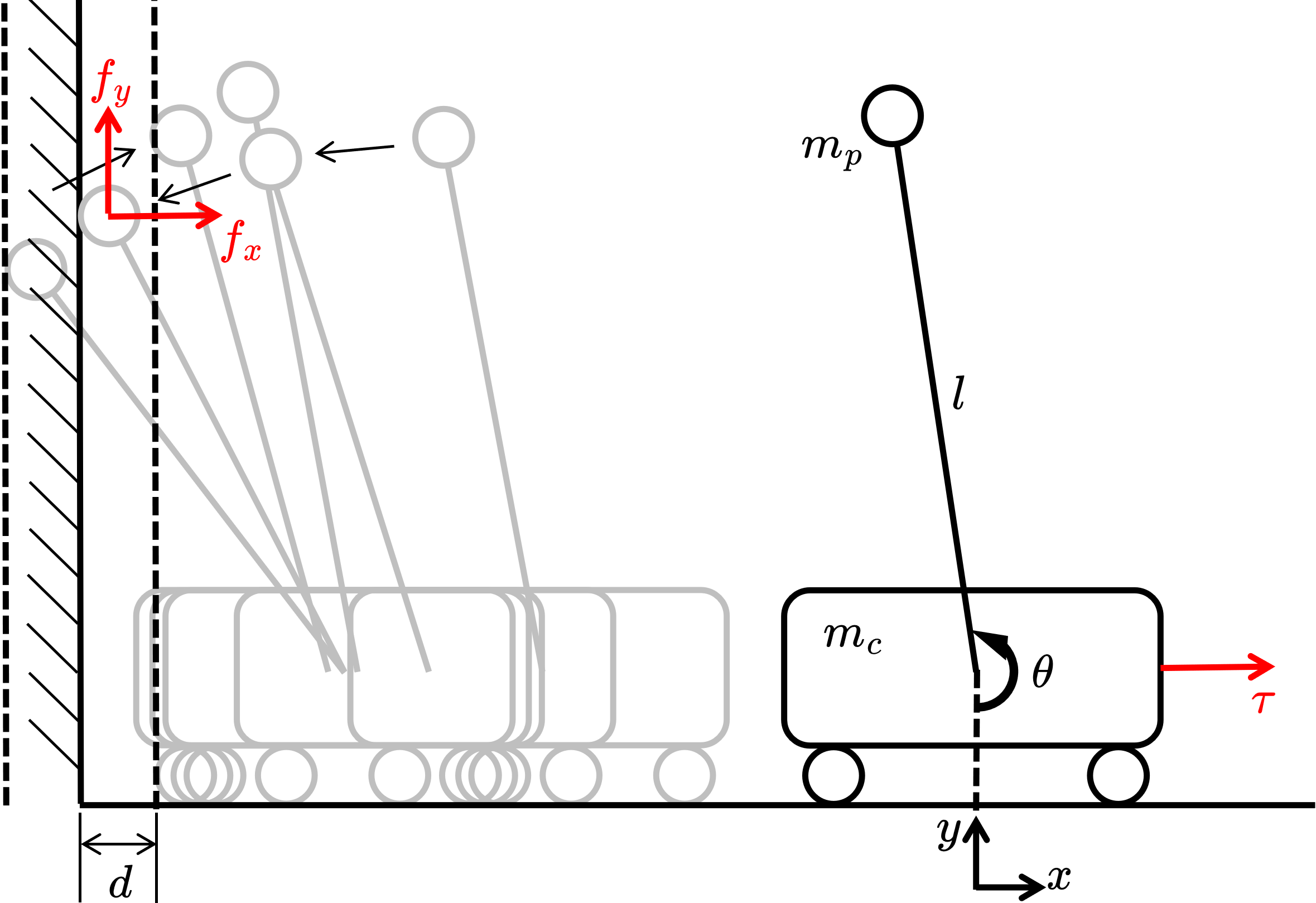}
	\caption{Illustration of the cart-pole system with wall contact. The cart-pole starts from an initially disturbed state and, due to limited control input, moves toward the wall to induce an impact that reverses the pole's velocity. After the impact, the system seeks to regain balance and return to the desired position. The wall position is uncertain within a range of $\pm d$.}
	\label{fig:cart_pole_system_with_wall}
\end{figure}
In this case study, a cart-pole system stabilizes itself under disturbances using contact with an uncertain wall (Fig.~\ref{fig:cart_pole_system_with_wall}). The system dynamics are given by:
\begin{equation}\label{eq:cart_pole_with_wall_dynamics}
\begin{aligned}
    M(\textbf{q})\ddot{\textbf{q}}+ H(\mathbf{q},\dot{\mathbf{q}}) = \begin{bmatrix}
    1 \\
    0
    \end{bmatrix}\tau + 
    J_c^{\top} \mathbf{F}
\end{aligned}
\end{equation}
where $\mathbf{q} = [x,\theta]^\top$ denotes the generalized coordinates, with $x$ the cart position and $\theta$ the pole angle. The cart and pole masses are $m_c=0.3\,\mathrm{kg}$ and $m_p=1\,\mathrm{kg}$, respectively, and the pole has length $l=0.4\,\mathrm{m}$ with mass concentrated at the tip. The system input is the external force $\tau$ acting on the cart.
$M(\mathbf{q})$ denotes the mass matrix, and $H(\mathbf{q},\dot{\mathbf{q}})$ captures Coriolis, centrifugal, and gravitational effects. The contact Jacobian at the pole tip is $J_c \in \mathbb{R}^{2\times 2}$, and the wall reaction force is $\mathbf{F} = [f_x, f_y]^\top \in \mathbb{R}^2$.

For the impact model, the wall is assumed to have a friction coefficient $\mu=0.7$, where
$|f_y| \leq \mu f_x$ and $\quad f_x > 0$.
The contact also has a restitution coefficient $e$, such that the pole-tip normal velocity relative to the wall before ($v_N^-$) and after ($v_N^+$) impact satisfies
    $v_N^+=-ev_N^-$.


\subsubsection{Cost and Constraints in the Nominal Formulation}
Let $\mathbf{x} = [\mathbf{q}^\top,\dot{\mathbf{q}}^\top]^\top = [x, \theta, \dot{x}, \dot{\theta}]^\top$ denote the system state. The running cost is defined as
\begin{equation}\label{eq:cartpolenominaltrajopt_runningcost}
    L_i = (\mathbf{x}_i-\mathbf{x}_{eq})^\top W_\mathbf{x}(\mathbf{x}_i-\mathbf{x}_{eq})\Delta t_i + w_\tau \tau_i^2\Delta t_i
\end{equation}
where $\mathbf{x}_{eq}$ denotes the equilibrium state, typically set to $\mathbf{x}_{eq} = [0, \pi, 0, 0]^\top$. The weighting matrices are chosen as $W_{\mathbf{x}} = \mathrm{diag}(10, 10, 1, 1)$ and $w_{\tau} = 1$, representing the relative importance of state and control costs, respectively. No terminal cost is applied in this formulation. The initial and terminal boundary conditions are set as 
$\mathbf{x}_0 = \mathbf{x}_{\mathrm{init}}$ and $\mathbf{x}_N = \mathbf{x}_{\mathrm{end}}$.

During free-motion phases, i.e., before and after contact, the system dynamics must satisfy:
\begin{subequations}
\begin{align}
\forall i \in [0, c) &\cup (c, N-1]: \nonumber\\
    &M(\mathbf{q}_i)\ddot{\mathbf{q}}_i + H(\mathbf{q}_i,\dot{\mathbf{q}}_i) = [1, 0]^\top \tau_i, \label{eq:cartpolenominaltrajopt_dynamics}\\
    &\mathbf{x}_{i+1} = \mathbf{x}_{i} +  \begin{bmatrix}
  \dot{\mathbf{q}}_i \\
  \ddot{\mathbf{q}}_i
  \end{bmatrix} \Delta t_i, \label{eq:cartpolenominaltrajopt_eulerintegration}\\
    &g(\mathbf{x}_i) = x_i + l \sin{\theta_i} - x_{\mathrm{wall}} > 0. \label{eq:cartpolenominaltrajopt_guard>0}
\end{align}
\end{subequations}
At contact ($i=c$), the system satisfies the guard condition:
\begin{equation}
    g(\mathbf{x}_i) = x_i + l \sin{\theta_i} - x_{\mathrm{wall}} = 0\label{eq:cartpolenominaltrajopt_guard=0}
\end{equation}
which triggers the corresponding state transition, implicitly defined by the following constraints:
\begin{subequations}
\begin{align}
&M(\mathbf{q}_i)\ddot{\mathbf{q}}_i + H(\mathbf{q}_i,\dot{\mathbf{q}}_i) = \begin{bmatrix}
    1 \\ 0
\end{bmatrix} \tau_i + J_c^\top \mathbf{F}, \, |f_y| \leq \mu f_x, \label{eq:cartpolenominaltrajopt_impactcons}\\
&\mathbf{x}_{i+1} = \mathbf{x}_{i} +  \begin{bmatrix}
  \mathbf{0} \\
  \ddot{\mathbf{q}}_i
  \end{bmatrix}\Delta t_{\mathrm{impact}}, \, J_{c,0} (\dot{\mathbf{q}}_{i+1} + e \dot{\mathbf{q}}_i) = 0.\label{eq:cartpolenominaltrajopt}
\end{align}
\label{eq:impactcombined}
\end{subequations}
Here,~\eqref{eq:cartpolenominaltrajopt_impactcons} define the dynamics and force constraints, where $\mathbf{F} = [f_x, f_y]^\top$ denotes the impact force acting on the pole tip, and $\Delta t_{\mathrm{impact}} = 0.001\,\mathrm{s}$ is the assumed impact duration (meaning the continuous forces act during the impact process, \cite{johnson2016hybrid}).  
The position and velocity transitions are enforced through~\eqref{eq:cartpolenominaltrajopt}. Note that $J_{c,0}$ denotes the first row of the contact Jacobian. 
Here, we assume that the position remains unchanged during the near-instantaneous contact event. Additional constraints are imposed to ensure that the left edge of the cart does not collide with the wall:
\begin{equation}
    \forall i \in [0, N]: \quad x_i - \tfrac{1}{2} w_{\mathrm{cart}} > x_{\mathrm{wall}}
\end{equation}
where $w_{\mathrm{cart}} = 0.08\,\mathrm{m}$ denotes the cart width.  

\subsubsection{Cost and Constraints in SURE}
In addition to the running cost on the common trajectory \eqref{eq:cartpolenominaltrajopt_runningcost}, 
the running cost for nodes on the rejoining branches is defined as
\begin{equation}
    L_{i,j}=(\mathbf{x}_{i,j}-\mathbf{x}_{eq})^\top W_\mathbf{x}(\mathbf{x}_{i,j}-\mathbf{x}_{eq})+w_\tau\tau_{i,j}^2
\end{equation}
where the same weighting factors are used as in the nominal case. The free-motion dynamics for all $i \in [0, \mathcal{K}_{e}) \cup (\mathcal{K}_{e}, N-1]$ follow 
\eqref{eq:cartpolenominaltrajopt_dynamics} and \eqref{eq:cartpolenominaltrajopt_eulerintegration}. When the system enters the branching phase, it must satisfy the following guard constraints:
\begin{subequations}
\begin{align}
&i=\mathcal{K}_0:\quad g(\mathbf{x}_i)=x_i+l\sin{\theta_i}-x_{\mathrm{wall}}=d,\\
&i=\mathcal{K}_{e}: \quad g(\mathbf{x}_i)=x_i+l\sin{\theta_i}-x_{\mathrm{wall}}=-d.
\end{align}
\end{subequations}
At $i = \mathcal{K}_0$, the pole tip reaches the right bound of the uncertainty range, 
while at $i = \mathcal{K}_{e}$, it reaches the left bound. For each node within the branching phase $i \in \mathcal{K}$, the state $\mathbf{x}_i$ is considered the pre-impact state, and a corresponding post-impact state $\mathbf{x}_{i,0}$ is defined implicitly by $\mathbf{x}_{i+1}$ in \eqref{eq:impactcombined}.
Each branch in the rejoining phase (i.e., $\forall i \in \mathcal{K}, j \in [0, N_r - 1]$) then evolves according to the free-body dynamics defined in \eqref{eq:cartpolenominaltrajopt_dynamics} and \eqref{eq:cartpolenominaltrajopt_eulerintegration}. Finally, all nodes beyond the branching and rejoining phases must remain outside the broadened guard region:
\begin{equation}
    \forall i \notin\mathcal{K}:\quad g(\mathbf{x}_i)=x_i+l\sin{\theta_i}-x_{\mathrm{wall}}>d
\end{equation}

\subsubsection{Results}
The trajectories from both nominal and SURE approaches are used as references to control the system in simulation. As the simulation platform, we developed a custom environment to reproduce the physical behavior of the cart-pole system and its interaction with the wall. For tracking the reference trajectory, a proportional-derivative (PD) controller with feedforward compensation is employed:
\begin{equation}
\tau = \mathbf{k}_p\big(\mathbf{q}_{des} - \mathbf{q}\big)
+ \mathbf{k}_d\big(\dot{\mathbf{q}}_{des} - \dot{\mathbf{q}}\big)
+ \tau_{des}
\end{equation}
where $\mathbf{k}_p\in\mathbb{R}^2$ and $\mathbf{k}_d\in\mathbb{R}^2$ are the proportional and derivative gain vectors of the PD controller, respectively. Their values are calculated using a Linear Quadratic Regulator (LQR) design. $\mathbf{q}_{des}(t)$, $\dot{\mathbf{q}}_{des}(t)$, and $\tau_{des}(t)$ denote the desired position, velocity, and control input obtained from the optimized trajectories.
For experiments with identical initial conditions, the cart–pole system employs identical PD controller gains.

We evaluate stabilization under two uncertainty sources: wall position (perception uncertainty) and restitution coefficient (model uncertainty). In each trial, both parameters are randomly sampled and fixed. The system is released from an (often aggressive) initial condition, and the controller tracks the reference trajectory while using wall contact for stabilization. A trial is successful only if:
\begin{enumerate}[label=(\roman*)]
\item the target state is reached within 10 seconds,
\item the pole tip contacts the wall at most once,
\item the pole does not fall, and
\item the cart does not penetrate the wall.
\end{enumerate}

\begin{figure}[btp]
	\centering
	\includegraphics[width=0.45\textwidth]{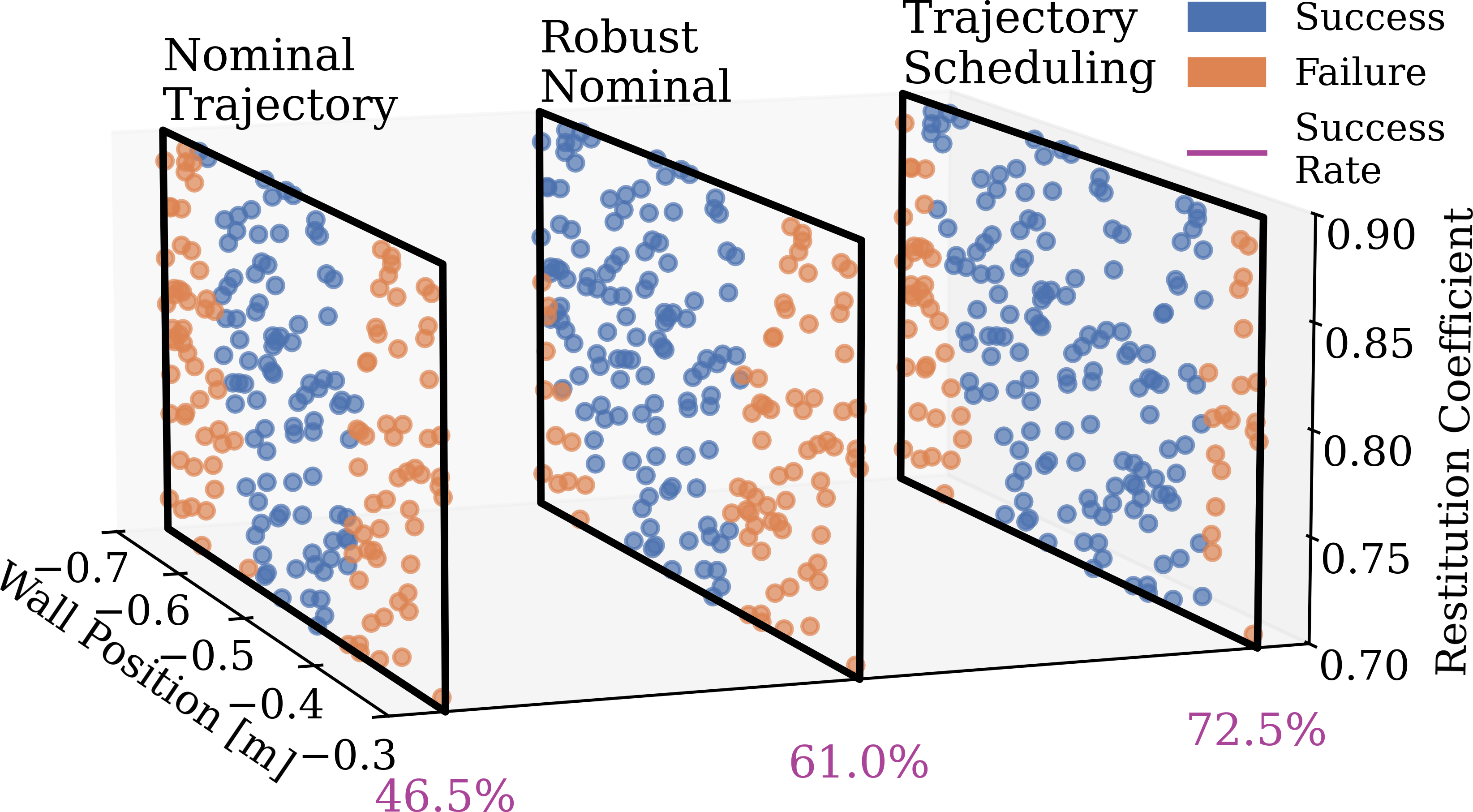}
	\caption{Robustness comparison under Initial Condition 4 for three reference trajectories. The nominal wall position is $x_{\mathrm{wall}}=-0.5\mathrm{m}$ with restitution coefficient 0.8. SURE uses 5 branches with uncertainty half-width $d=0.05\mathrm{m}$. The wall position varies in $[-0.7,-0.3]\mathrm{m}$ and the restitution coefficient in $[0.7,0.9]$. We sample 200 points in this uncertainty space and evaluate success rates for all three approaches.}\label{fig:robustness_comp_nominal_robust}
\end{figure}

Four sets of initial conditions are evaluated. Each case starts from a different initial state but shares the same terminal state, $\mathbf{x}_{\mathrm{end}} = [0, \pi, 0, 0]^\top$. For each case, we compare three reference trajectories: the nominal trajectory, the robust nominal trajectory (Section~\ref{sec:rejoining_nominal_trajectory}), and the scheduled robust trajectories (Section~\ref{sec:trajectory_scheduling}).
The results are summarized in Table~\ref{table:initial_and_terminal_conditions}, and representative successful and failed samples for Condition~4 are shown in Fig.~\ref{fig:robustness_comp_nominal_robust}.

\begin{table}[tb]
\vspace*{5pt}
\centering
\setlength{\tabcolsep}{3pt}
\caption{Simulation Results for the Cart–Pole System}
\label{table:initial_and_terminal_conditions}
\begin{tabular}{@{}ccccc|ccc@{}} 
\toprule
& \multicolumn{4}{c}{\textbf{State}} & \multicolumn{3}{c}{\textbf{Success Rate}} \\
\midrule
\makecell{\textbf{Initial}\\\textbf{Condition}} & \makecell{$x$\\$[m]$} & \makecell{$\dot{x}$\\$[m/s]$} & \makecell{$\theta$\\$[rad]$} & \makecell{$\dot{\theta}$\\$[rad/s]$} & Nominal &\makecell{Robust\\Nominal}&\makecell{Trajectory\\Scheduling}\\
\midrule
1 & 0 & 0 & $\pi$ & 5.5 & 58.5\% & 60.0\% & 75.0\% \\
2 & 0 & 0 & $\pi$ & 6.5 & 37.0\% & 43.0\% & 48.0\% \\
3 & 0 & -1.0 & 3.53 & 3.5 & 37.0\% & 57.0\% & 70.0\%  \\
4 & 0 & -0.5 & 3.45 & 4.5 & 46.5\% & 61.0\% & 72.5\% \\
\midrule
\textbf{Total}
& & & &
& 44.8\% & 55.3\% & 66.4\% \\
\bottomrule
\end{tabular}
\end{table}

First, we observe that the nominal trajectories already exhibit a certain degree of robustness to variations in wall position and restitution coefficient with a success rate of 44.8\%, even though these uncertainties are not explicitly considered during optimization.
In comparison, the SURE solution with trajectory scheduling demonstrates a substantially higher success rate of 66.4\%---even when evaluated over a wider uncertainty range than that specified during optimization (the intended range being $e = 0.8$, $x_\mathrm{wall} = -0.5 \pm 0.05\,\mathrm{m}$).

To further investigate whether the trajectory itself---independent of the scheduler---provides improved robustness, we compare the nominal trajectory with the robust nominal trajectory. In both cases, the controller is constrained to track a single deterministic reference. The results show that, even without a scheduler to accommodate different contact timings, the robust nominal trajectory exhibits an improvement in success rate (55.3\%) compared to the nominal trajectory.

\begin{figure}[btp]
    \vspace*{5pt}
	\centering
	\includegraphics[width=0.48\textwidth]{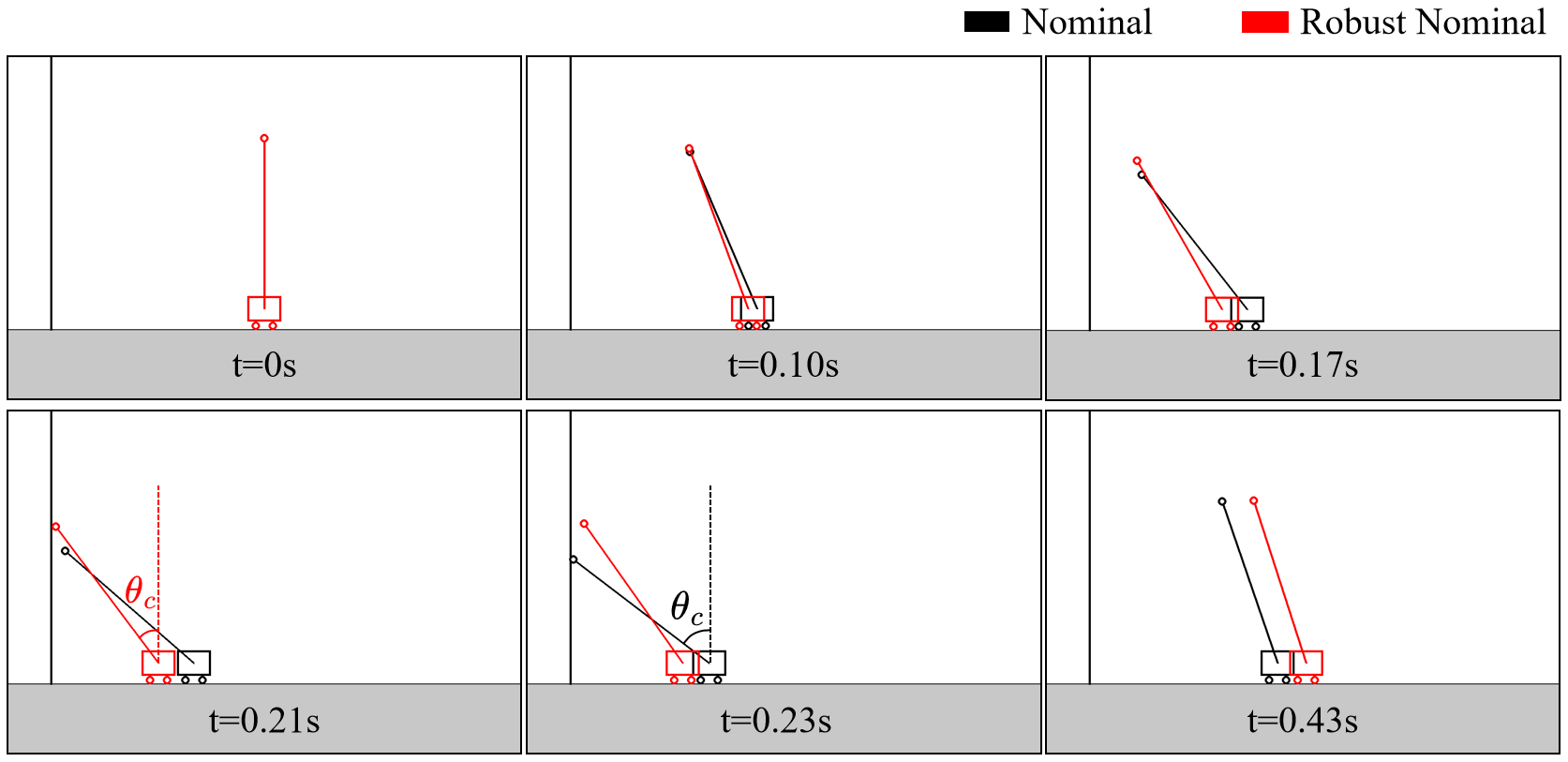}
	\caption{Qualitative comparison of nominal and robust nominal trajectories in simulation, with $\mathbf{x}_{\mathrm{init}} = [0, \pi, 0, 5.5]^\top$, $\mathbf{x}_{\mathrm{end}} = [0, \pi, 0, 0]^\top$, nominal wall position $x_\mathrm{wall} = -0.5\text{m}$, and $d = 0.05\text{m}$. $\theta_c$ denotes the pole angle at contact. Despite sharing the same nominal wall position, the trajectories differ in both wall approach and balance recovery strategies, leading to different post-impact stability.}
	\label{fig:qualitative_explanation}
\end{figure}

This suggests that the introduction of branching during TO not only modifies the motion locally around the contact event but also reshapes the pre- and post-contact segments. As a result, the system adopts a more effective global motion strategy. This effect is illustrated in Fig.~\ref{fig:qualitative_explanation}. In the nominal solution, the system knows the exact wall position; thus, the pole is allowed to swing freely as it approaches the wall. Around $t = 0.23\,\mathrm{s}$, just before contact, the pole angle $\theta$ becomes excessively large, causing the pole to nearly collapse. The system then relies on a strong elastic rebound from the wall to recover its balance. In contrast, in the robust nominal solution, the cart begins moving leftward in coordination with the pole, preventing the pole angle from growing too large. At $t = 0.21\,\mathrm{s}$, when contact occurs, the pole remains within a much smaller angular range, requiring only a mild impact to restore balance. Consequently, even if contact occurs slightly later than expected, the system can still maintain stability. After impact, the system shifts rightward to regain balance. Notably, the robust nominal solution exhibits a longer buffering distance, allowing additional room to compensate for discrepancies in contact timing or other uncertainties. These adjustments in the motion strategy explain the enhanced robustness of the robust nominal trajectory.

\subsubsection{Trade-off Between Optimality and Computing Time}
\begin{figure}[!t]
    \vspace*{5pt}
	\centering
	\includegraphics[width=0.48\textwidth]{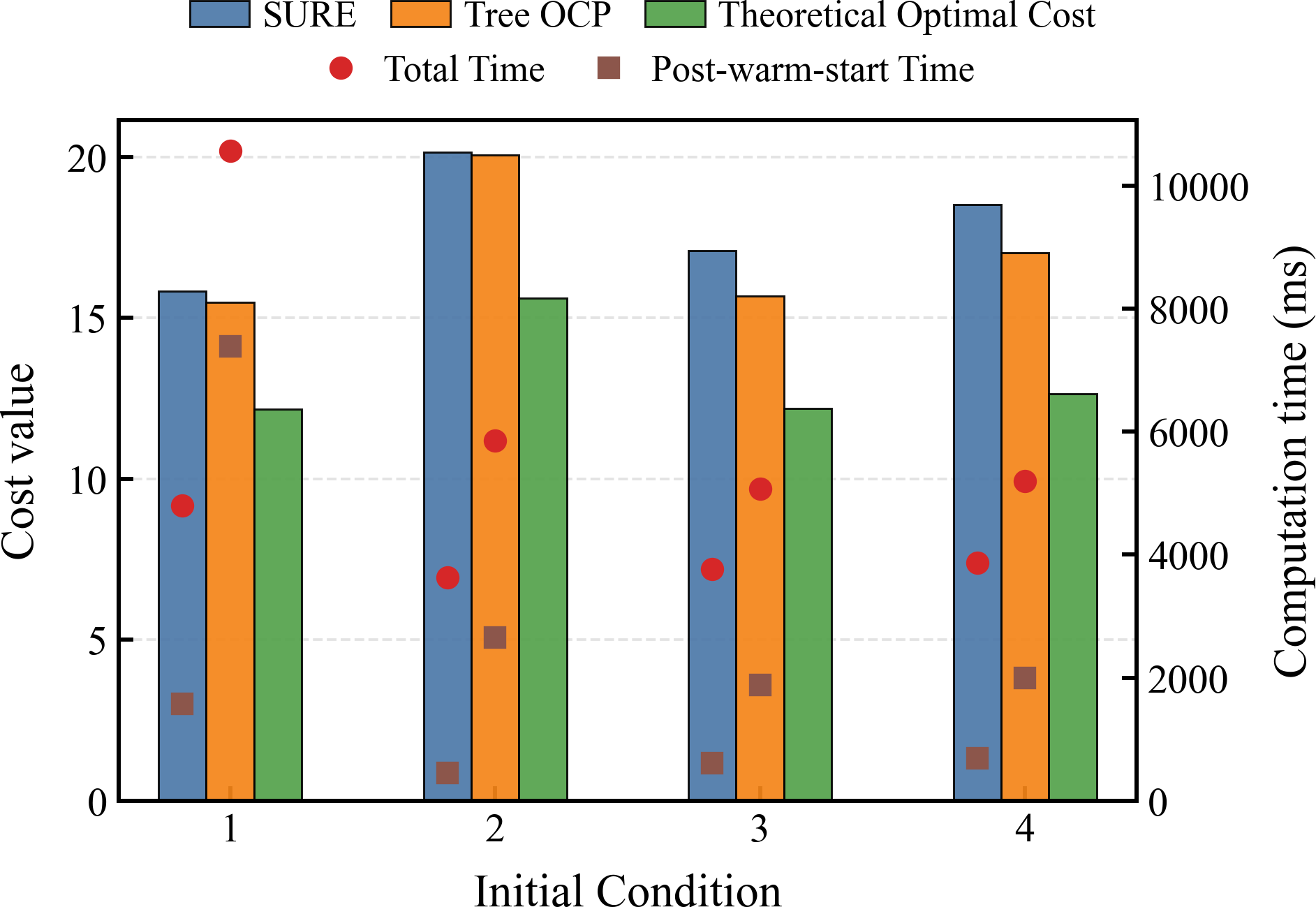}
	\caption{Optimality gap vs computation time trade-off under the four initial conditions listed in Table~\ref{table:initial_and_terminal_conditions}. For both SURE and Tree OCP configurations, the total number of post-impact nodes (branches plus common final trajectory) is fixed at 100. SURE corresponds to the configuration used in this paper, with $N_r = 7$ nodes per branch and 93 nodes for the common final trajectory. Tree OCP corresponds to the formulation in Fig.~\ref{fig:tree_ocp_qp}, where branches do not rejoin and each uses $N_r = 100$ nodes to reach the terminal state. Results show that increasing $N_r$ does not significantly reduce cost, while computation time increases sharply.}\label{fig:optimality_computing_trade_off}
\end{figure}

In addition to evaluating robustness, we also examine the computational cost of the proposed method. Figure~\ref{fig:optimality_computing_trade_off} compares the optimized cost and computation time under the four initial conditions.

Since not all configurations converge, we first solve a simplified SURE formulation that is guaranteed to be feasible. The resulting solution is then used as an initial guess for the full configuration. Reported computation times include both total time and post-warm-start time. All experiments are run on a laptop with Ubuntu 24.04, an Intel Core Ultra 9 185H CPU, and 32 GB RAM.

From the solution of SURE, we compute the actual wall position associated with each branch based on its pre-impact state. Using each inferred wall position as a deterministic wall location, we then solve a nominal TO problem. This yields a nominal cost value for each branch; taking the average of these values gives an average cost, shown as the green bar in Fig.~\ref{fig:optimality_computing_trade_off}. This represents the theoretical optimal cost for this problem with perfect knowledge.

Compared to Tree OCP, enforcing a common final trajectory across all branches introduces additional constraints and thus increases cost. However, the results show that this loss in optimality is small, while the reduction in computation time is substantial: On average, the cost increases by only 4.85\%, whereas total computation time decreases by 39.78\% and the post-warm-start time decreases by 76.05\%.

\subsection{Case Study II: Ball Catching}
\label{sec:casestudy2}
As shown in Fig.~\ref{fig:egg_catching_demo}, we consider a Unitree Z1 robotic manipulator with a lightweight container as its end effector. The manipulator dynamics are given by:
\begin{equation}
M(\textbf{q})\ddot{\textbf{q}}+H(\mathbf{q},\dot{\mathbf{q}})=\boldsymbol{\tau}
\end{equation}
where $\mathbf{q}\in\mathbb{R}^{6}$ denotes the joint angles, $\mathbf{M}(\mathbf{q})\in\mathbb{R}^{6\times 6}$ is the mass matrix, and $\mathbf{H}(\mathbf{q},\dot{\mathbf{q}})\in\mathbb{R}^{6}$ collects the Coriolis, centrifugal, and gravitational terms. $\boldsymbol{\tau}\in\mathbb{R}^{6}$ is the vector of joint torques. A ball falls freely from an initial height $h_{0}$ with uncertainty $\pm d$ and no horizontal velocity. Given a prescribed release time, the manipulator aims to catch the ball while minimizing impact and move it to a final position.

\subsubsection{Cost and Constraints in the Nominal Formulation}
Assuming the ball mass is small relative to the robot’s effective mass, impact minimization can be approximated by minimizing the relative velocity between the ball and end effector at contact. We further assume that the impact does not significantly affect the robot trajectory. Let $\mathbf{x}=[\mathbf{q}^\top,\dot{\mathbf{q}}^\top]^\top$ denote the system state; the nominal TO cost consists of two terms:
\begin{equation}
    \min_{\mathbf{x},\ddot{\mathbf{q}},\boldsymbol{\tau}\Delta t} \quad\left\|\mathbf{v}_{\mathrm{ball},c} -\mathbf{v}_{\mathrm{ee},c}\right\|^2 + w_{a}\sum_{i=0}^{N-1}\ddot{\mathbf{q}}_i^2\Delta t_i
\end{equation}
The first term penalizes the squared relative velocity between the ball and the end effector at the contact node $c$, where $\mathbf{v}_{\mathrm{ball},c}\in\mathbb{R}^3$ and $\mathbf{v}_{\mathrm{ee},c}\in\mathbb{R}^3$ denote their Cartesian velocities. The latter is given by $\mathbf{v}_{\mathrm{ee},c}=J_t(\mathbf{q}_c)\dot{\mathbf{q}}_c$, where $J_t(\mathbf{q}_i)\in\mathbb{R}^{3\times 6}$ denotes the translational Jacobian. The second term penalizes the accumulated joint accelerations to promote smooth motion, with $w_a$ denoting the corresponding weight. For all nodes except the last, we require $\forall i \in [0,N-1]:$
\begin{subequations}
\allowdisplaybreaks
\begin{align}
    &\mathbf{M}(\mathbf{q}_i)\ddot{\mathbf{q}}_i
  + \mathbf{H}(\mathbf{q}_i,\dot{\mathbf{q}}_i)
  = \boldsymbol{\tau}_i, \,
    \mathbf{x}_{i+1} = \mathbf{x}_i +
  \begin{bmatrix}
  \dot{\mathbf{q}}_i\\
  \ddot{\mathbf{q}}_i
  \end{bmatrix}
  \Delta t_i, \label{eq:dynamics_integration}\\
    &1-\mathbf{R}_{\mathrm{ee},i}[0,0,1]^\top < \varepsilon, \,(\mathbf{p}_{\mathrm{ee},i}-\mathbf{p}_{\mathrm{ball},0})_{xy}\|^2 < \varepsilon.\label{eq:objectcatchtrajopt}
\end{align}
\end{subequations}
Here,~\eqref{eq:dynamics_integration} enforces the manipulator dynamics and forward-Euler state update. To ensure reliable catching, \eqref{eq:objectcatchtrajopt} keeps the container horizontal, and constrains the end-effector center to remain aligned with the ball's vertical trajectory. $\mathbf{p}_{\mathrm{ball},0}$ denotes the ball's initial position, and $(\cdot)_{xy}$ denotes projection onto the $x$-$y$ plane.  
$\mathbf{p}_{\mathrm{ee},i}$ 
and $\mathbf{R}_{\mathrm{ee},i}$ 
are the translational and rotational components of the end effector pose, respectively, computed via forward kinematics ($\varepsilon=10^{-3}$). The guard condition is defined as the vertical separation between the ball and the end-effector bottom. Prior to contact, the ball must remain above the end effector:
\begin{equation}
\forall i \in [0,c):\quad g(\mathbf{x}_i) = p_{\mathrm{ball},i,z} - r_{\mathrm{ball}} - p_{\mathrm{ee},i,z} > 0
\label{eq:noearlycontact}
\end{equation}
where $p_{(\cdot),z}$ denotes the $z$-component of position and $r_{\mathrm{ball}}$ is the ball radius. At contact, the guard becomes zero and the contact event is triggered:
\begin{equation}
i=c:\quad g(\mathbf{x}_i) = p_{\mathrm{ball},i,z} - r_{\mathrm{ball}} - p_{\mathrm{ee},i,z} = 0
\end{equation}
The ball is assumed to remain attached to the end effector after contact. The pre-impact ball kinematics are given by
\begin{subequations}
\begin{align}
&\mathbf{v}_{\mathrm{ball},i} = \mathbf{v}_{\mathrm{ball},0} + [0,0,-g\,t_i]^\top, \\
&\mathbf{p}_{\mathrm{ball},i} = \mathbf{p}_{\mathrm{ball},0} + \mathbf{v}_{\mathrm{ball},0} t_i + \left[0,0,-\tfrac{1}{2} g\, t_i^2\right]^\top.
\end{align}
\end{subequations}
where $\mathbf{v}_{\mathrm{ball},0}$ is the initial ball velocity (typically zero) and $t_i$ is the accumulated time at node $i$, $t_i = t_0 + \sum_{k=0}^{i-1} \Delta t_k$.


\subsubsection{Cost and Constraints in SURE}
The free-motion dynamics for all $i\in[0,\mathcal{K}_{e})\cup(\mathcal{K}_{e},N-1]$ follow \eqref{eq:objectcatchtrajopt}. As the ball's initial height is uncertain, the formulation must account for a range of possible contact times. The earliest contact corresponds to the lowest initial height, whereas the latest contact corresponds to the highest initial height; all other feasible contact times lie between these two extremes. Over the branching phase, the following constraints are imposed:
\begin{subequations}
\begin{align}
i=\mathcal{K}_{0}:&g(\mathbf{x}_i) = p_{\mathrm{ball},i,z} - r_{\mathrm{ball}} - p_{\mathrm{ee},i,z}=d, \label{eq:objectcatchtrajopt_firstbranchingnode}\\
i=\mathcal{K}_{e}: & g(\mathbf{x}_i) = p_{\mathrm{ball},i,z} - r_{\mathrm{ball}} - p_{\mathrm{ee},i,z}=-d, \label{eq:objectcatchtrajopt_lastbranchingnode}\\
\forall i \in \mathcal{K}: &\|\mathbf{v}_{\mathrm{ee},i} -\mathbf{v}_{\mathrm{ball},i}\|^2\le v^2_{\lim}.\label{eq:objectcatchtrajopt_vdiffbound}
\end{align}
\end{subequations}
Here,~\eqref{eq:objectcatchtrajopt_firstbranchingnode} and~\eqref{eq:objectcatchtrajopt_lastbranchingnode} enforce the guard condition at the first and last nodes of the branching phase, respectively. At each branching node, \eqref{eq:objectcatchtrajopt_vdiffbound} bounds the relative velocity between the ball and the end effector, where $v_{\lim}>0$ denotes the upper bound on the allowable relative speed. The cost function in the robust optimization problem then minimizes this bound together with the accumulated joint accelerations:
\begin{equation}
\min_{\mathbf{x},\dot{\mathbf{x}},\Delta t,v_{\lim}}  v_{\lim} + w_{a}\sum_{i=0}^{N-1}\ddot{\mathbf{q}}_i^2\Delta t_i + w_{a}\sum_{i\in\mathcal{K}}\sum_{j=0}^{N_r-1}\ddot{\mathbf{q}}_{i,j}^2\Delta t_{i,j}
\end{equation}
For each branching node $i\in\mathcal{K}$, we assume state continuity at contact ($\mathbf{x}_{i,0}=\mathbf{x}_i$) and enforce that the corresponding branch rejoins the common final trajectory at its final node.
Each branch evolves according to \eqref{eq:objectcatchtrajopt}.
Prior to the branching phase, we enforce that contact cannot occur prematurely, \eqref{eq:noearlycontact}.

\subsubsection{Results}
\begin{figure}[tbp]
    \centering
    \subfloat[Position]{
        \includegraphics[width=0.22\textwidth]{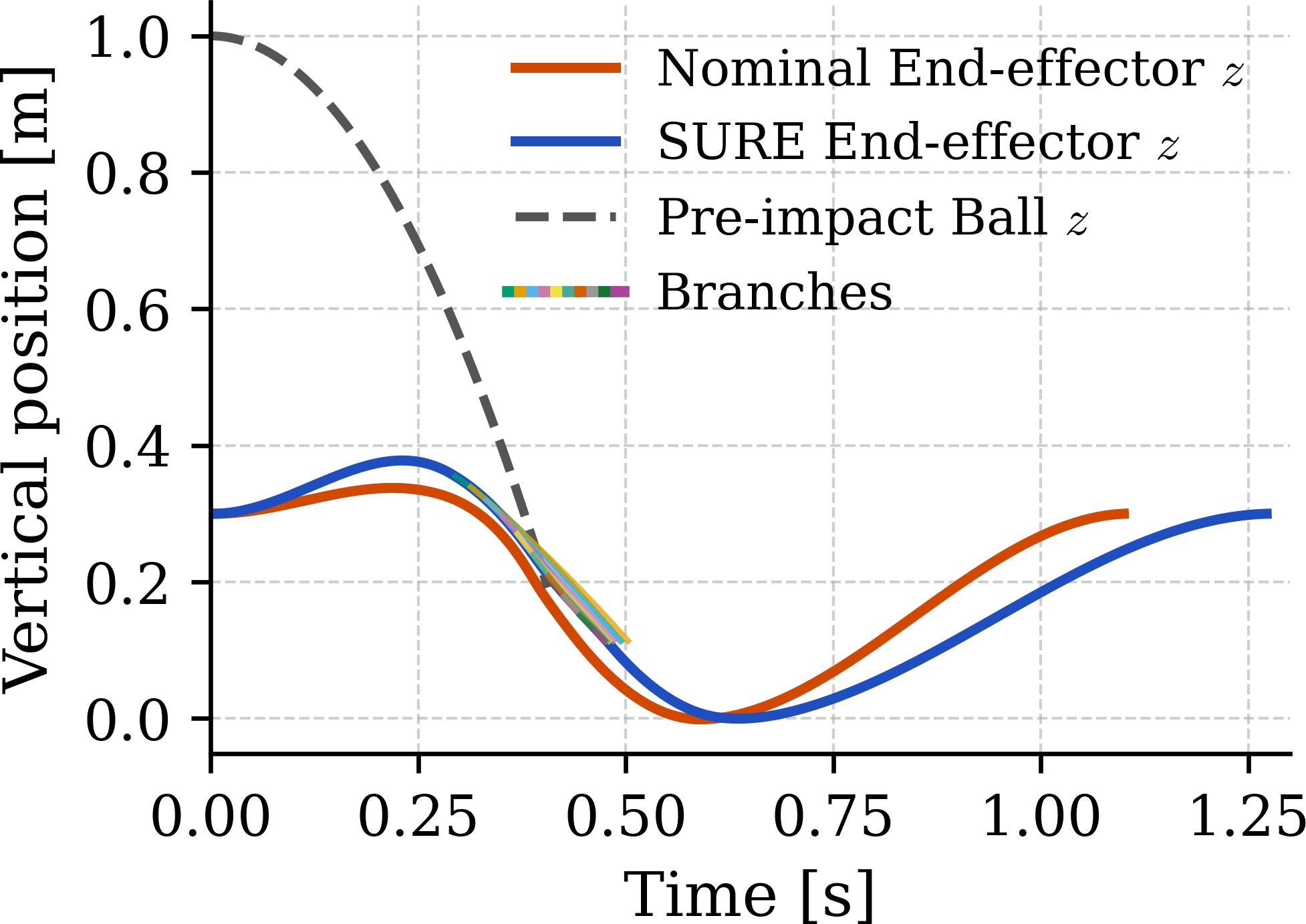}
        \label{fig:position_nominal_and_robust_nominal_with_branches}
    }\hfill
    \subfloat[Velocity]{
        \includegraphics[width=0.22\textwidth]{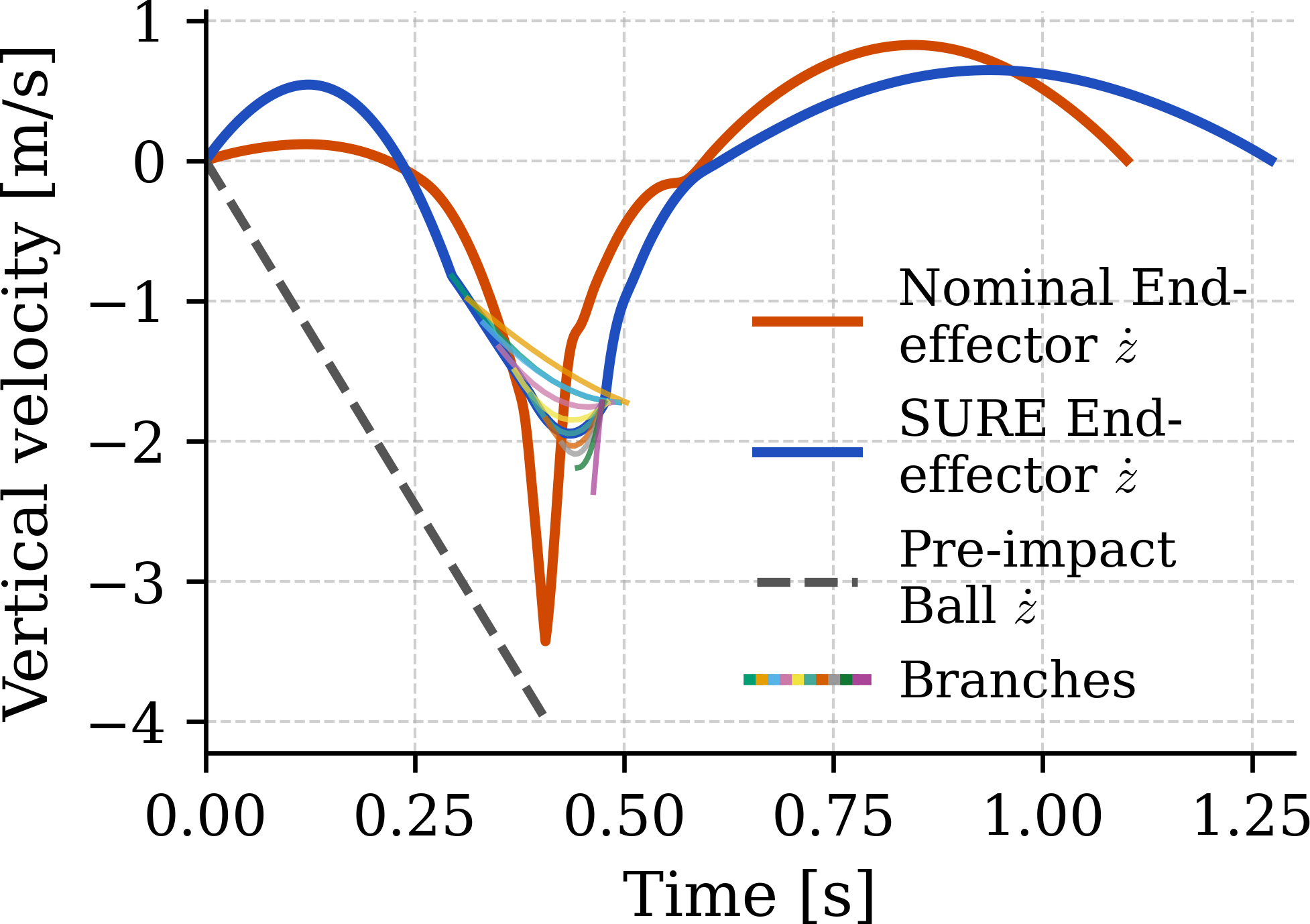}
        \label{fig:velocity_nominal_and_robust_nominal_with_branches}
    }
    \caption{Nominal trajectory: the end effector moves slightly upward, then descends rapidly assuming deterministic contact timing. SURE robust nominal trajectory: the end effector rises higher, enabling longer co-travel with the egg, and descends more smoothly to accommodate timing uncertainty.}\label{fig:object_catching_solution_comparison}
\end{figure}

Figure~\ref{fig:object_catching_solution_comparison} compares trajectories from the nominal and SURE formulations, including position and velocity profiles. The end effector moves from $\mathbf{p}{\mathrm{ee},0}=[0,0,0.3]\,\mathrm{m}$, catches the ball, and returns to $\mathbf{p}{\mathrm{ee},e}=[0,0,0.3]\,\mathrm{m}$, while the ball is released from $\mathbf{p}_{\mathrm{ball},0}=[0,0,1.0]\,\mathrm{m}$. In SURE, ten branches are used with an initial-height uncertainty half-width of $d=0.20\,\mathrm{m}$.

\begin{figure}[tbp]
    \centering
    \subfloat[Nominal]{
        \includegraphics[width=0.47\linewidth]{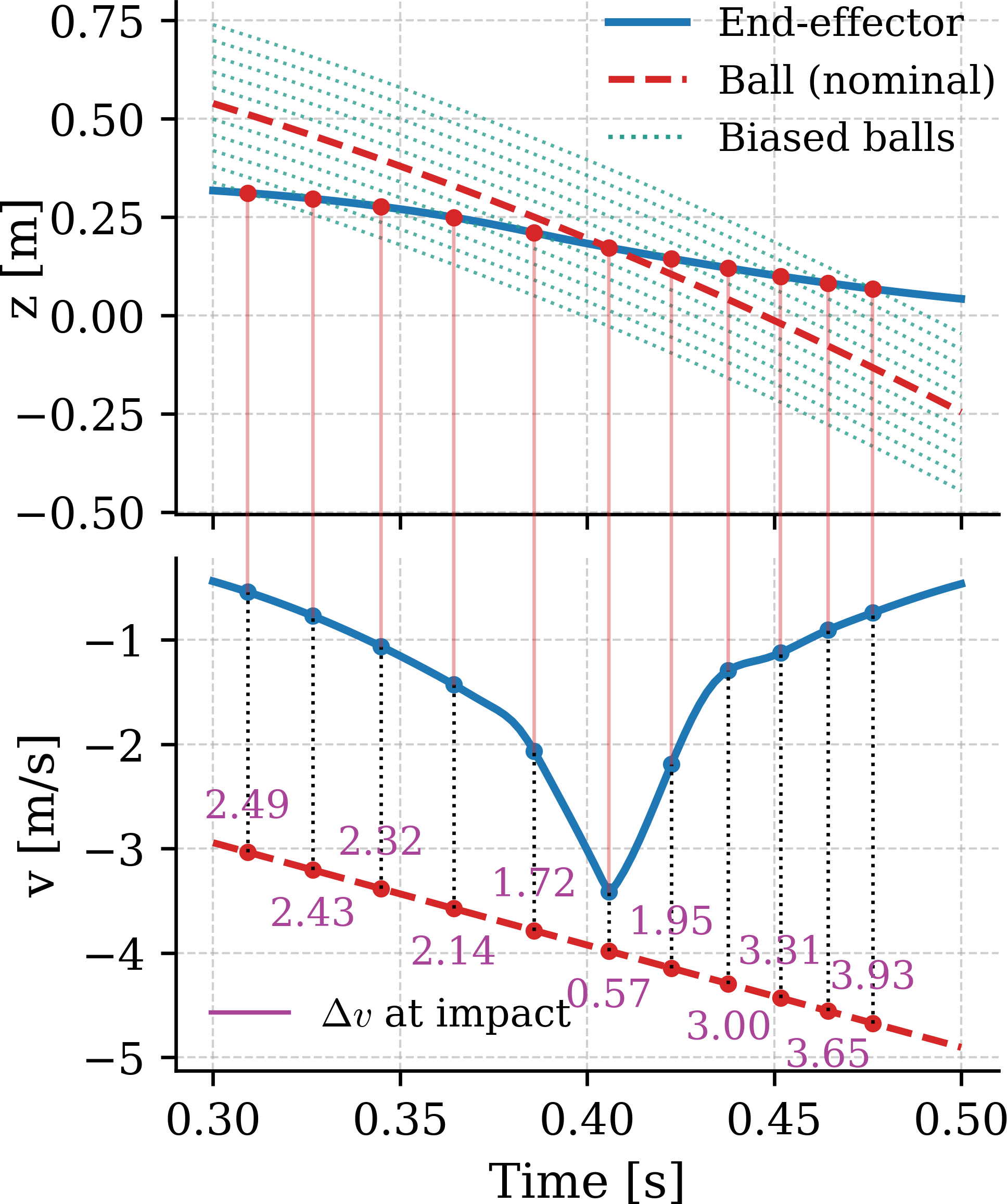}
        \label{fig:nominal_impact_analysis}
    }
    \subfloat[Robust Nominal]{
        \includegraphics[width=0.47\linewidth]{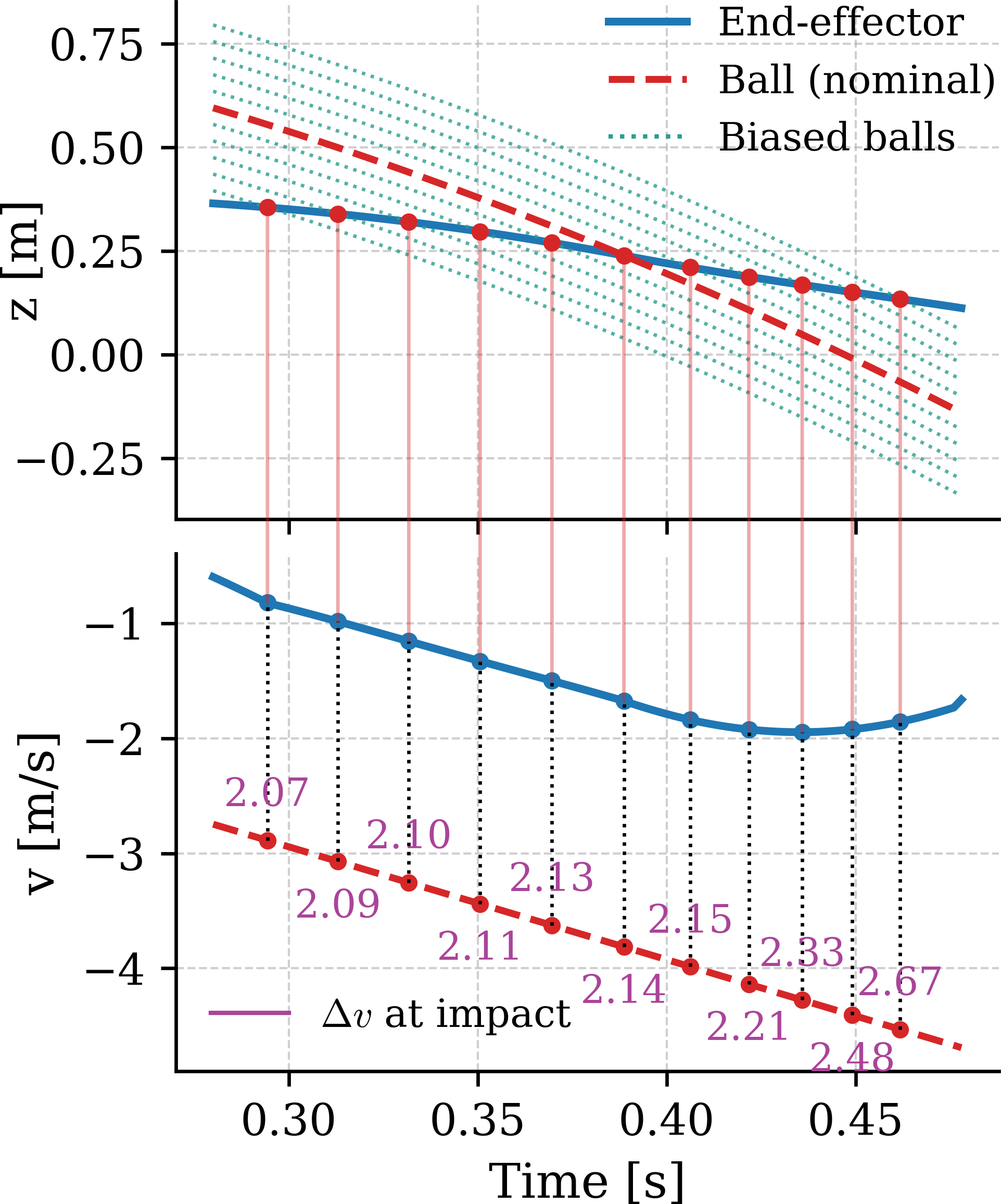}
        \label{fig:robust_nominal_impact_analysis}
    }
    \caption{Relative velocity between the ball and end effector for varying initial ball heights. The ball is released from 11 uniformly spaced heights in $[p_{\mathrm{ball},0,z}-0.2\,\mathrm{m}, p_{\mathrm{ball},0,z}+0.2\,\mathrm{m}]$. The top row shows vertical positions, and the bottom row shows velocities. Vertical dashed lines mark contact times, and purple annotations indicate relative impact velocity. (a) Nominal trajectory, with a maximum impact velocity of $3.93\mathrm{m/s}$. (b) Robust nominal trajectory, reducing the maximum to $2.67\mathrm{m/s}$.}
    \label{fig:theoretical_analysis_relative_velocity}
\end{figure}

To analyze impact robustness, we evaluate the relative velocity between the ball and end effector at contact for varying initial ball heights (Fig.~\ref{fig:theoretical_analysis_relative_velocity}). The results show that the robust nominal trajectory provides significantly improved robustness. The nominal trajectory performs well only at the nominal height; any deviation increases impact velocity except for the nominal and two nearby cases. In contrast, the robust nominal trajectory keeps the relative velocity $v_{\lim}$ below 2.67m/s across all cases, effectively limiting impact under timing uncertainty. The nominal trajectory reaches up to 3.93 m/s, a 47\% increase.

\subsubsection{Real-world Experiments}
We conduct physical experiments in which the robot catches an egg using either the nominal or SURE robust nominal trajectory (Fig.~\ref{fig:object_catching_solution_comparison}) under identical initial conditions. Trajectory scheduling is not implemented due to the lack of reliable contact sensing.
As shown in Fig.~\ref{fig:egg_catching_demo}, the egg is released by a Franka Emika Panda mounted on a higher table, while the Unitree Z1 robot on a lower table performs the catching motion. At the start of each trial, the Z1 moves to an initial configuration where the end effector is directly beneath the egg. Both robots are connected to a laptop running Ubuntu 20.04 with a real-time kernel and remain idle until a trigger signal is issued. Upon triggering, the Panda opens its gripper to release the egg, and the Z1 executes the corresponding tracking trajectory.

\begin{table}[t]
\vspace*{5pt}
\centering
\setlength{\tabcolsep}{3pt}
\caption{Result of the egg drop experiment}
\label{table:egg_drop_experiment_result}
\begin{tabular}{@{}ccccccccccccc
@{}} 
\toprule
$t_{\mathrm{wait}}$ & Trajectory & 1 & 2 & 3 & 4 & 5 & 6 & 7 & 8 & 9 & 10& \makecell{Success\\Rate}\\
\midrule
\multirow{2}{*}{\makecell{0.125s}} &nominal& \ding{55} & \ding{51} & \ding{51} & $\circ$ & $\circ$ & \ding{55} & $\circ$& \ding{55} & \ding{55} & \ding{55} &35\%\\
&SURE & \ding{51} & \ding{51} & \ding{51} & \ding{51} & \ding{51} & \ding{51} & \ding{51} & $\circ$ & \ding{51} & \ding{51} & 95\%\\
\multirow{2}{*}{\makecell{0.118s}} & nominal & \ding{55} & \ding{55} & \ding{51} & \ding{51} & \ding{55} & \ding{51} & \ding{51} & $\circ$ & \ding{51} & \ding{55} &55\%\\
& SURE & \ding{51}  & $\circ$ & $\circ$ & \ding{55} & $\circ$ & \ding{51} & \ding{51} & \ding{51} & \ding{51} & \ding{51} &75\%\\
\midrule
\multirow{2}{*}{\makecell{\textbf{Total}}} &nominal & & & &&&&&&&&45\%\\
 &SURE &&&&&&&&&&&85\%\\
\bottomrule
\end{tabular}
\end{table}

Opening the Panda gripper requires a non-negligible time, so the Z1 uses a fixed delay $t_{\mathrm{wait}}\approx 0.12\mathrm{s}$ for approximate synchronization. Ideally, this aligns the egg release with the start of the catching motion. However, perfect synchronization is not achievable due to gripper actuation delay, communication and processing latency, and controller discretization. We exploit this inherent timing uncertainty to evaluate the robustness of the proposed trajectories.

For each trajectory, we tune $t_{\mathrm{wait}}$ so that the average contact time matches the nominal one. This is done by iteratively adjusting $t_{\mathrm{wait}}$ and running ten drop-catch trials until early and late contacts are balanced. The calibrated values are $t_{\mathrm{wait}}=0.118\mathrm{s}$ (nominal) and $t_{\mathrm{wait}}=0.125\mathrm{s}$ (SURE). We then perform ten additional trials per setting and classify outcomes as:
\begin{enumerate}[label=(\roman*)]
\item successful catch without damage (\ding{51}, score $+1$),
\item contact followed by upward bounce and re-capture without break ($\circ$, score $+0.5$),
\item failure due to breakage or drop (\ding{55}, score $0$).
\end{enumerate}
Results are summarized in Table~\ref{table:egg_drop_experiment_result}. Under both timing settings, SURE consistently outperforms the nominal trajectory, achieving an overall success rate of 85\% versus 45\%, corresponding to a 40\% improvement under timing uncertainty.

\section{Conclusion} 
\label{sec:conclusion}
In this paper, we proposed an efficient formulation for robust trajectory optimization that can handle uncertainties in contact timing. The main novelty of our formulation lies in branching possible solutions at different contact timings and rejoining all solutions at a common point outside the uncertain region. Through extensive simulation and experimental demonstrations, we showed that our framework adapts its strategy in the presence of uncertainty, resulting in more robust solutions than a nominal formulation at a lower computational cost than other robust approaches.

In future work, we plan to extend SURE to multiple contact events. The multi-contact formulation is obtained by concatenating multiple branching, rejoining, and common final trajectory phases. This relies on the assumption that consecutive contacts are sufficiently separated in time so that each branching–rejoining phase can be treated independently. Furthermore, we are interested to learn uncertainty-conditioned policies from our robust formulation under various uncertainty conditions. We are also interested in extending the framework to floating-base systems, with the goal of enabling robust loco-manipulation behaviors.




\bibliographystyle{IEEEtran}
\bibliography{IEEEabrv,references}

\end{document}